\relax
\documentclass[letterpaper]{article} 
\usepackage{aaai22}  
\usepackage{times}  
\usepackage{helvet}  
\usepackage{courier}  
\usepackage[hyphens]{url}  
\usepackage{graphicx} 
\usepackage{natbib}  
\usepackage{caption} 
\DeclareCaptionStyle{ruled}{labelfont=normalfont,labelsep=colon,strut=off} 
\usepackage{algorithm}
\usepackage{algorithmic}
\usepackage{amssymb}
\usepackage{multirow}
\usepackage{booktabs}
\usepackage{amsmath}
\usepackage{newfloat}
\usepackage{listings}
\floatstyle{ruled}
\newfloat{listing}{tb}{lst}{}
\floatname{listing}{Listing}
\nocopyright
\title{S+PAGE: A Speaker and Position-Aware Graph Neural Network Model for Emotion Recognition in Conversation}
\author{
    Chen Liang, Chong Yang, Jing Xu, Junyang Huang, Yongliang Wang, Yang Dong
}

\begin{document}

\maketitle

\begin{abstract}

Emotion recognition in conversation (ERC) has attracted much attention in recent years for its necessity in widespread applications. Existing ERC methods mostly model the self and inter-speaker context separately, posing a major issue for lacking enough interaction between them. In this paper, we propose a novel Speaker and Position-Aware Graph neural network model for ERC (S+PAGE), which contains three stages to combine the benefits of both Transformer and relational graph convolution network (R-GCN) for better contextual modeling. Firstly, a two-stream conversational Transformer is presented to extract the coarse self and inter-speaker contextual features for each utterance. Then, a speaker and position-aware conversation graph is constructed, and we propose an enhanced R-GCN model, called PAG, to refine the coarse features guided by a relative positional encoding. Finally, both of the features from the former two stages are input into a conditional random field layer to model the emotion transfer. 
Extensive experiments demonstrate that our model achieves  state-of-the-art performance on three ERC datasets.
\end{abstract}

\section{1 Introduction}
Emotion recognition in conversation (ERC), which aims to identify the emotion of each utterance in a conversation, is a task arousing increasing interests in many fields. With the prevalence of social media and intelligent assistants, ERC has great potential applications in several areas, such as emotional chatbots, sentiment analysis of comments in social media and healthcare intelligence, for understanding emotions in the conversation with emotion dynamics and generating emotionally coherent responses. 
ERC remains a challenge. Both lexicon-based \cite{wu2006emotion,shaheen2014emotion} and deep learning-based \cite{colnerivc2018emotion} text emotion recognition methods that treat each utterance individually fail in this task as these works ignore some conversation-specific characteristics.

\begin{figure}[t]
\centering
\includegraphics[width=0.85\columnwidth]{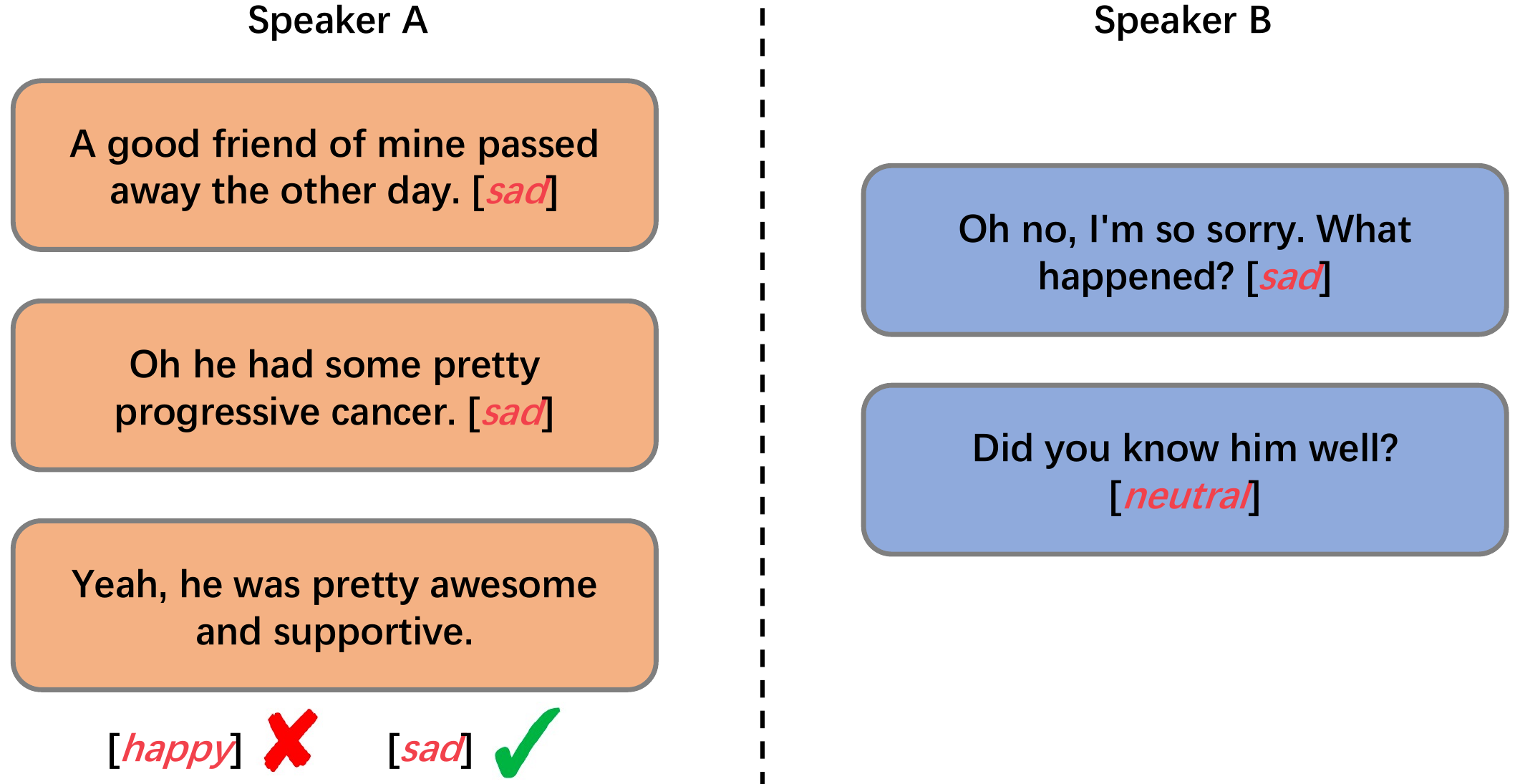} 
\caption{A dialogue from IEMOPCAP, in which the emotion of the last utterance by speaker A will be wrongly classified if the dialogue context is not taken into consideration.}
\label{fig1}
\end{figure}

Empirical results show that conversational context plays an important role in the ERC task \cite{poria2019emotion}.
As demonstrated in Figure \ref{fig1}, the third utterance by speaker A will be assigned a wrong emotion label if the history conversation information is blind to the model.  
In the past few years, recurrent neural network (RNN)-based solutions, such as CMN \cite{cmn}, ICON \cite{icon} and DialogueRNN \cite{dialoguernn}, have dominated this field due to the sequential nature of conversational context. 
Nonetheless, they share some inherent limitations: 1) RNN model performs poorly in grasping distant contextual information; 2) RNN-based methods are not capable of handling large-scale multiparty conversations. 


With the rise of graph neural network (GNN) \cite{wu2020comprehensive} in many natural language processing (NLP) tasks, researchers pay increasing attention to GNN-based ERC methods recently. Instead of modeling only sequential data recurrently in RNN, GNN is designed to capture all kinds of graph structure information via various aggregation algorithms. Existing GNN-based ERC methods, such as DialogueGCN \cite{dialoguegcn}, RGAT \cite{rgat} and DAG-ERC \cite{dag}, which are the state of the art, have demonstrated the superiority of GNN in modeling conversational context. A directed graph is constructed on each dialogue in these methods, where the nodes denote the individual utterances, and the edges indicate relationships between utterances. However, we notice that these methods do not work well on modeling speaker-specific context, which is also important in the ERC task. For example, in Figure \ref{fig1} the third utterance spoken by speaker A is more influenced by speaker A's prior utterances rather than the second utterance spoken by speaker B, even though the latter is closer. 
Thus, in contextual modeling, we should consider both the emotional influence that speakers have on themselves during a conversation, i.e., self-speaker context, and context on the entire conversation flow, i.e., inter-speaker context, as well as the interaction between them.

On the other hand, none of the existing methods actually exploit the fine-grained temporal information, i.e., the relative distance between utterances, because the original graph aggregation algorithms do not propagate distance-related message. We surmise that the relative distance is helpful to grasp more temporal information, and thus can further improve the performance.

In this paper, we propose a novel \textbf{S}peaker and \textbf{P}osition-\textbf{A}ware \textbf{G}NN model for \textbf{E}RC (S+PAGE) to settle the above drawbacks of exisiting methods. Our model contains three stages to fully consider both self and inter-speaker context.
Specifically, given a sequence of utterances in the same dialogue, we first leverage a \textbf{T}wo-\textbf{S}tream \textbf{C}onversational \textbf{T}ransformer (TSCT) with the attentive masking mechanism to get both the same speaker's and the whole dialogue's contextual features. Then, guided by the speaker identity and utterance order, we construct a speaker and position-aware conversation graph. The  \textbf{P}osition-\textbf{A}ware \textbf{G}NN model (PAG), which is an enhanced relational graph convolution network (R-GCN), is utilized to refine the contextual features with self and inter-speaker dependency. Particularly, we introduce relational relative positional encoding in the aggregation algorithm, in order to make PAG capable of capturing fine-grained temporal information. Finally, the global transfer of emotion labels is modeled by a conditional random field (CRF) layer with the features from both TSCT and PAG. 
Experimental results demonstrate the superiority of our model compared with state-of-the-art models. Ablation study illustrates the effectiveness of the proposed Transformer and position-aware graph structure in modeling the conversational context. To conclude, our contributions are as follows:

\begin{itemize}
    \item We propose a novel graph neural network-based ERC method, called S+PAGE, to incorporate all kinds of conversational context information in the model at the same time.
    \item We present a two-stream conversational Transformer architecture to extract both conversational and speaker-specific context-aware features, which is also capable of handling multiparty conversations on large scale with high efficiency.
    \item We apply a relational relative positional encoding on the graph neural network to capture fine-grained temporal information in a conversation.
    \item We conduct extensive experiments on several ERC datasets, which demonstrate that our proposed model can significantly promote the performance.
\end{itemize}

\section{2 Related Works}
\subsection{2.1 Emotion Recognition in Conversation} 
Emotion recognition in conversation is a popular area in NLP. Many ERC datasets have been scripted and annotated in the past few years, such as IEMOCAP \cite{iemocap}, MELD \cite{meld}, DailyDialog \cite{dailydialog}, EmotionLines \cite{emotionlines} and EmoryNLP \cite{emorynlp}. 
IEMOCAP, MELD, and EmoryNLP are multimodal datasets, containing acoustic, visual and textual information, while the remaining two datasets are textual.

In recent years, ERC solutions are mostly deep learning-based models. CMN \cite{cmn} and ICON \cite{icon} utilize gated recurrent unit (GRU) and memory networks to capture the dialogue dynamics. In IAAN \cite{iaan} and DialgueRNN \cite{dialoguernn}, attention mechanisms are applied to interact between the party state and global state. With the rise of Transformer and graph neural networks in NLP tasks, many works have also introduce them into the ERC task. \citet{ket} propose KET, which is a structure of hierarchical Transformers assisted by external commonsense knowledge. DialogueXL \cite{dialogxl} applies dialogue-aware self-attention to deal with the multi-party structures. In DialogueGCN \cite{dialoguegcn} and RGAT \cite{rgat}, GCN \cite{kipf} and GAT \cite{gat} are applied to refine the features with speaker dependencies and temporal information. DAG-ERC \cite{dag} applies a directed acyclic graph for conversation representation and it achieves the state-of-the-art performance on multiple ERC datasets.

\subsection{2.2 Transformer} 
\citet{transformer} first propose Transformer for machine translation task, whose success subsequently has been proved in various down-stream NLP tasks. Self-attention mechanisms endow Transformer with the ability of capturing longer-range dependency among elements of an input sequence than the RNN structure.
\citet{longformer} propose a novel self-attention mechanism for feature extraction of long documents.
Pre-trained models such as BERT \cite{bert} and GPT \cite{gpt} use Transformer encoder and decoder respectively to learn representations on large-scale datasets. 
In our model, we present a modified Transformer structure to encode the contextual information in a conversation.

\begin{figure*}[t]
\centering
\includegraphics[width=0.9\textwidth]{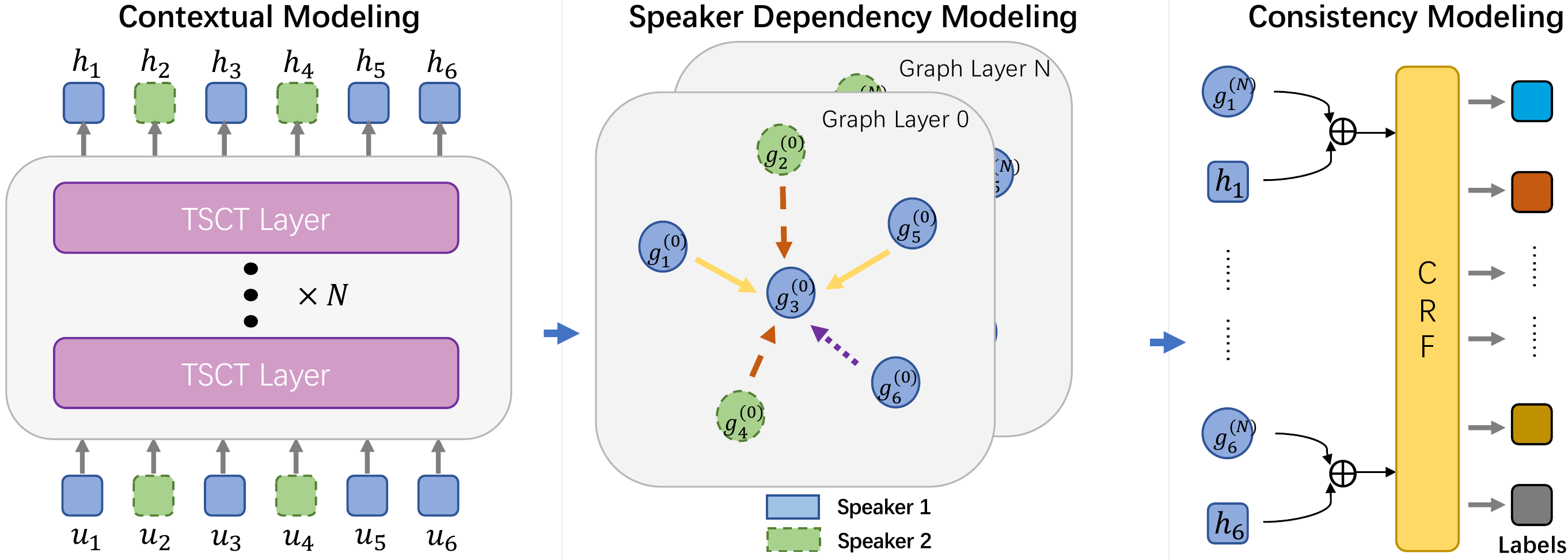} 
\caption{The framework of S+PAGE.}
\label{fig2}
\end{figure*}

\subsection{2.3 Graph Neural Network} 
Graph neural network has attracted a lot of attention in recent years, which learns a target node’s representation by
propagating neighbor information in the graph.
\citet{kipf} propose a simple and well-behaved layer-wise propagation rule for neural network models and demonstrate its effectiveness in semi-supervised classification tasks.
Better aggregation methods for large graphs are proposed in GAT \cite{gat} and GraphSage \cite{graphsage}.
\citet{rgcn} propose R-GCN to deal with the highly multi-relational data characteristic by assigning different aggregation structures for each relation type. 

\section{3 Methodology}



The framework of our model is shown in Figure 2. We decompose the emotion classification procedure into three stages, i.e., contextual modeling, speaker dependency modeling, and global consistency modeling. In the first stage, we present a novel conversation-specific Transformer to get coarse contextual features as well as cues of speaker information. Then, a graph neural network is proposed to refine the features with speaker dependency and temporal information guided by relational relative positional features. Subsequently, we employ conditional random field to model the context of global emotion consistency, i.e., the emotion transfer. The interaction between self and inter-speaker context is involved in both of the former two stages.

\subsection{3.1 Problem Definition}
The ERC task is to predict emotion labels (e.g., Happy, Sad, Neutral, Angry, Excited, and Frustrated) for utterances \(u_1; u_2; \cdots ; u_{N}\), where N denotes the number of utterances in a conversation. 
Let $S$ be the number of speakers in a given dataset. $P$ is a mapping function, and $p_{s} = P(u_i)$ denotes utterance $u_i$ uttered by speaker $p_{s}$, where $s \in \{1, \cdots, S\}$.

\subsection{3.2 Utterance Encoding} \label{section:utterance}
Following previous works \cite{dialoguegcn,dialoguernn}, 
we use a simple architecture consisting of a single convolutional layer followed by a max-pooling layer and a fully connected layer to extract context-independent textual features of each utterance. The input of this network is the 300 dimensional pre-trained 840B GloVe vectors \cite{glove}. 
We use the output features, denoted as $\vec{u_{i}}$, as the representation of each utterance.

\begin{figure*}[ht]
\centering
\includegraphics[width=0.9\textwidth]{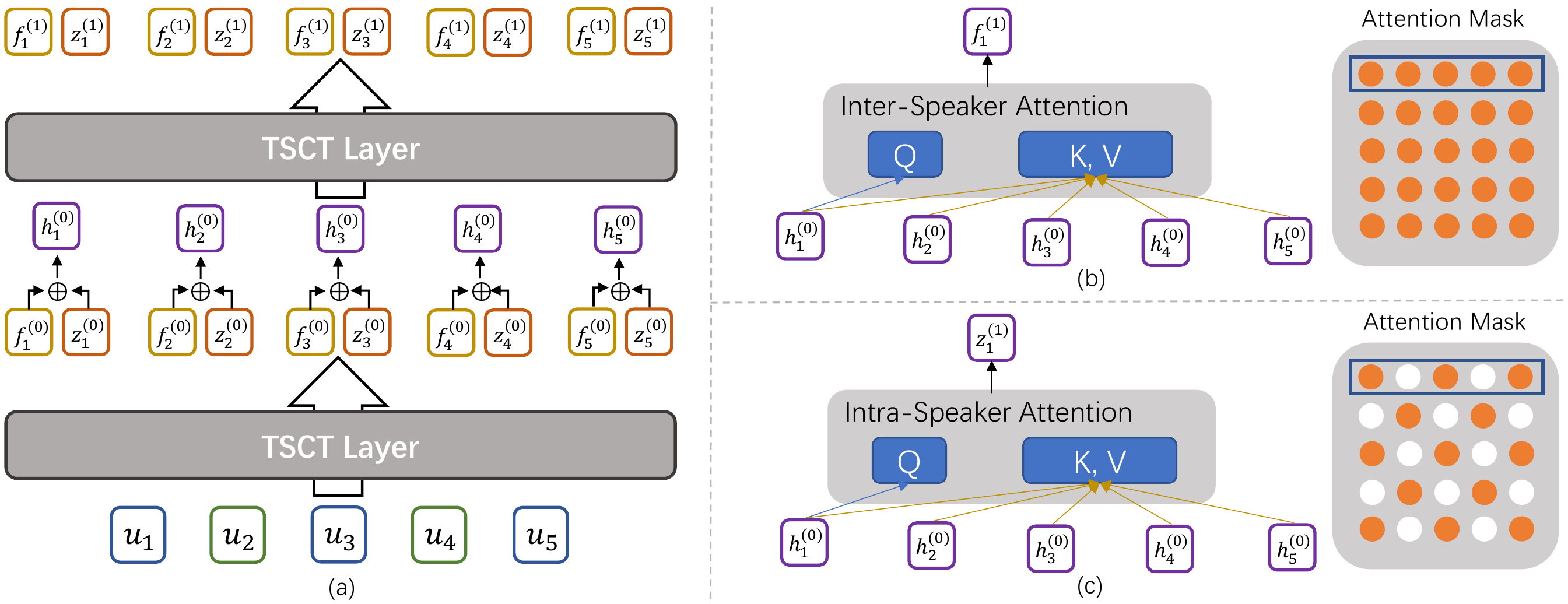} 
\caption{(a) Structure of the two-stream conversational Transformer, (b) Inter-speaker self-attention, (c) Intra-speaker self-attention }
\label{fig3}
\end{figure*}

\subsection{3.3 Contextual Modeling}
\subsubsection{Two-Stream Conversational Transformer}
We present a \textbf{T}wo-\textbf{S}tream \textbf{C}onversational \textbf{T}ransformer (TSCT) to better extract the contextual representation of each utterance in a conversation, which is also capable of handling multi-party conversations efficiently.
The structure of TSCT is shown in Figure \ref{fig3}. The collection of utterance representations $U = {\vec{u_1}; \vec{u_2}; \cdots ; \vec{u_{N}}} $ is taken as the input. We design a multi-head self-attention mechanism, composed of two streams, i.e., the inter-speaker self-attention stream and the intra-speaker self-attention stream.

\subsubsection{Inter-Speaker Self-Attention} The inter-speaker self-attention is the same as the self-attention in vanilla Transformer, in which each utterance can attend to all positions in the dialogue as shown in Figure \ref{fig3}(b). It is calculated as: 

\begin{equation}
    q_{i}^{l}, k_{i}^{l}, v_{i}^{l} = h_{i}^{l-1}W_{iq}^{l}, h_{i}^{l-1}W_{ik}^{l},  h_{i}^{l-1}W_{iv}^{l}
\end{equation}
\begin{equation}
    f_{i}^{l} = softmax(\frac {q_{i}^{l}(k_{i}^{l})^{T}} {\sqrt{d}}) v_{i}^{l}
\end{equation}

where $W_{iq}^{l}$, $W_{ik}^{l}$ and $W_{iv}^{l}$ are three learnable weight matrices for attention head $i$ at layer $l$. 

\subsubsection{Intra-Speaker Self-Attention} The intra-speaker self-attention models speaker-specific contextual information by only computing attention on the same speaker's utterances in a dialogue. In this way, the model is able to capture the emotional  influence  that  speakers  have  on  themselves during the conversation. It is implemented by the attentive masking strategy as illustrated in Figure \ref{fig3}(c) and formulated as:

\begin{equation}
    z_{i}^{l} = softmax(\frac {q_{i}^{l}(k_{i}^{l})^{T}} {\sqrt{d}} + m) v_{i}^{l}
\end{equation}

where $m \in \mathbb{R}^{N \times N}$ is the attentive masking matrix. The elements of $m$ are set as below: 

\begin{equation}
m_{ij} =\left\{
\begin{array}{rcl}
-\infty   &      & P(u_{i}) \neq P(u_{j})\\
0         &      & otherwise\\
\end{array} \right. 
\end{equation}

where $P(\cdot)$ is the function that maps the utterance and its corresponding speaker.

Each attention head $i$ of the $l$-th layer in TSCT, denoted as $head_{i}^{l}$, is the concatenation of the $f_{i}$ and $z_{i}$, and the output of the multi-head attention can be formulated as follows: 

\begin{equation}
    MultiHead_{i}^{l} = \|_{i=1}^{M} head_{i}^{l} \\
\end{equation}
where $\|$ denotes concatenation operation. $M$ is the number of attention heads, while $1 \leq i \leq M$.


Following the structure of the original Transformer, the output of the TSCT layer can be generated by passing $MultiHead_{i}^{l}$ through a normalization layer followed by a feed-forward network:
\begin{equation}
h^{l} = \text{LayerNorm}(\text{FeedForward}(MultiHead_{i}^{l}))
\end{equation}

\subsection{3.4 Speaker Dependency Modeling}

After extracting the contextual features, we construct a conversation graph based on the speaker identity and the order of utterances. We design a novel position-aware graph neural network to capture both speaker dependency and temporal information in a dialogue by introducing relative positional encoding. 

\subsubsection{Graph Architecture}
We construct a directed graph, $\mathcal{G} = (\mathcal{V},\mathcal{E}, \mathcal{R}, \mathcal{W}) $, for each dialogue with $N$ utterances. The nodes in the graph are the utterances in the conversation, i.e., $V = \{ u_{1}; u_{2}; \cdots, u_{N} \}$. $(v_{i}, v_{j}, r_{ij}) \in \mathcal{E}$ denotes a labeled edge (relation), where $r_{ij} \in \mathcal{R}$ is a relation type, defined according to speaker identity and relative distance. $\mathcal{W}$ represents the set of edge weights.

\subsubsection{Nodes}
 Feature vector $g_{i}$ of each node $v_{i}$ is initialized as the output of the TSCT layer, i.e., $h_{i}$. Feature vector $g_{i}$ is modified by the aggregation algorithm through the stacked graphical layers in GNN.
 The output feature is described as $g_{i}^{l}$, where $l$ denotes the number of layers.

\subsubsection{Edges}
Instead of only focusing on past utterances, we take converse influence into account \cite{dialoguegcn}. 
We construct edges $\mathcal{E}$ with a sliding window for each utterance. The window sizes $p$ and $f$ denote the number of past and future utterances from the target utterance. Each utterance node $v_{i}$ has an edge  with $p$ utterances of the past: $v_{i-1},v_{i-2},...,v_{i-p}$ ,  $f$ utterances of the future: $v_{i+1},v_{i+2},...,v_{i+f}$, and itself.

\subsubsection{Edge Types} 
\begin{figure}[t]
\centering
\includegraphics[width=0.75\columnwidth]{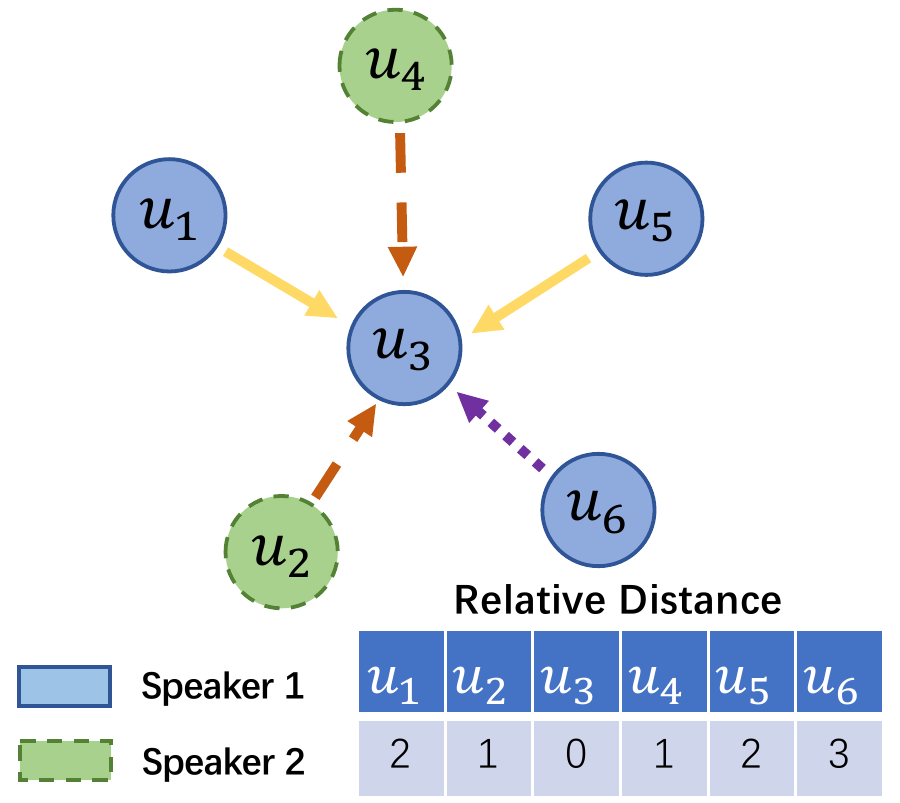} 
\caption{An example of dialogue graph construction. Different types of arrows denote different edge types. The table shows the relative distances for each utterance to the target utterance $u_{3}$.}
\label{fig4}
\end{figure}

The relation type $r \in R$ is determined by the \textit{speaker identity} and \textit{relative distance}. The maximum relative distance is clipped to the window sizes. For example, if we set both future and past window sizes to $n$, there are $n$ relative distances. Thus, assuming there are $m$ distinct speakers in a dialogue, there should be $N_{e} = m * n$ relation types in the constructed graph $\mathcal{G}$.
Two utterances share the same edge type only if they are uttered by the same speaker and have equal distances from the target utterance.
For example, in Figure \ref{fig4} $u_{1}$ and $u_{5}$ belong to the same edge type, while $u_{1}$ and $u_{6}$ belong to different edge types.

\subsubsection{Edge Weights} Edge weights are computed by an attention mechanism. 
The particular attentional setup in our model closely follows the work of GAT \cite{gat}. The input of the attention module is a set of node features, i.e., $g = {\vec{g_{1}}, \vec{g_{2}}, ..., \vec{g_{N}}} \in \mathbb{R}^{F}$, where $F$ is the dimension of each node feature. Motivated by \cite{shaw}, which shows that absolute positional encoding is not effective for the model to capture the information of relative word order, we inject relative positional encoding into the attention mechanism. Fully expanded out, the edge weights, denoted as $\alpha_{ij} \in \mathcal{W}$, computed by the position-aware graph attention mechanism can be described as:

\begin{equation}
    \alpha_{i j}=\frac{\exp L_{ij}}{\sum_{k \in N i} \exp L_{ik}}
    \label{ewight}
\end{equation} 

\begin{equation}
    L_{ij} = LeakyReLU\left(\vec{a}^{T}\left[W g_{i} \| (W g_{j}+\beta_{i j})\right]\right)
\end{equation}

where $W \in \mathbb{R}^{F^{'} \times F}$ is a weight matrix applied to every node, and $F^{'}$ is the dimension of $\beta_{ij}$.   
$N_{i}$ is the number of nodes linked with node $i$.
$a \in \mathbb{R}^{2F^{'}}$ is a parametrized weight vector. $\cdot^{T}$ represents transposition, and $\|$ is the concatenation operation. $\beta_{ij}$ denotes the relative position representation between utterance $i$ and utterance $j$ in a dialogue, which is encoded by a learnable embedding matrix $E_{p}$:

\begin{equation}
 \beta_{ij} = E_{p}(o(u_{j}) - o(u_{i}))   
\end{equation} 

where $o(\cdot)$ is a mapping function between utterance and its absolute position in the dialogue sequence. Notice that we use a signed relative position instead of the relative distance in definition of edge type to keep the temporal order. Thus, we have $E_{p} \in \mathbb{R}^{w * F^{'}}$, where $w = p + f$ denotes the whole window size.

\subsubsection{PAG} 
We introduce a novel graph neural network to propagate utterance features, named \textbf{P}osition-\textbf{A}ware \textbf{G}raph neural network (PAG). 
Inspired by R-GCN \cite{rgcn}, we define the following aggregation algorithm to calculate the forward-pass update of a node in the graph:
\begin{equation}
    g_{i}^{l}=\sigma\left(\sum_{r \in \mathcal{R}} \sum_{j \in N_{i}^{r}} \frac{\alpha_{i j}}{c_{i, r}} W_{r}^{l} g_{i}^{l-1}+\alpha_{i i} W_{o}^{l} g_{i}^{l-1}\right)
\end{equation}
where $g_{i}^{l}$ is the hidden state of node $i$ in the $l$-th layer. $N_{i}^{r}$ denotes the set of neighbors of utterance $i$ under the edge type $r \in R$. $c_{i,r}$ is a normalization constant, and we set $c_{i,r} = |N_{i}^{r}|$ in our experiment. $W_r^{l}$ and $W_o^{l}$ are learnable weight matrices, and $\sigma(\cdot)$ is an activation function. Different from R-GCN, we use edge weights calculated by Equation \ref{ewight} to involve fine-grained temporal information in a conversation.

Furthermore, between the stacked layers of PAG, we employ a fusion function, which is formulated as:

\begin{equation}
    {g}_{i}^{l} = Fuse(g_{i}^{l}, g_{i}^{l-1} )
\end{equation}

where $l \ge 1$. The fusion function is designed as a gated sum of two features:

\begin{equation}
    Fuse(\mathbf{a}, \mathbf{b})=\mathbf{z} * \mathbf{a}+(1-\mathbf{z}) * \mathbf{b}
\end{equation}
\begin{equation}
    \mathbf{z}=sigmoid\left(\mathbf{W}_{z}[\mathbf{a} ; \mathbf{b} ; \mathbf{a} * \mathbf{b} ; \mathbf{a}-\mathbf{b}]+\mathbf{b}_{z}\right)
\end{equation}
where $\mathbf{a}$ and $\mathbf{b}$ denote the feature vectors, and $\mathbf{W}_{z}$ is a trainable weight matrix. 

\subsection{3.5 Consistency Modeling}

\subsubsection{Conditional Random Field}
To model the label consistency, i.e., the emotion transfer on a dialogue, a linear chain conditional random field is employed to yield final emotion tags of each utterance. Following the description of \cite{lample}, for an input set of utterances $U = {u_{1}, u_{2}, ...,u_{n}}$ and a sequence of tag predictions $y = {y_{1}, y_{2},..,y_{n}}$, $y_{i} \in {1, \cdots ,K}$ (K is the number of emotion tags), the score of the sequence is defined as:

\begin{equation}
    s(\mathbf{U}, \mathbf{y})=\sum_{i=0}^{n} T_{y_{i}, y_{i+1}}+\sum_{i=1}^{n} Q_{i, y_{i}}
\end{equation}
 
where $T \in \mathbb{R}^{K \times K}$ is the trainable transition matrix of the labels, and $Q \in \mathbb{R}^{n \times K}$ denotes the emmision matrix. In our model, $P$ is computed by the concatenation of the last two modules' output features, following a linear layer and a softmax function. The model is trained to maximize the log-probability of the correct tag sequence:

\begin{equation}
    \log (p(\mathbf{y} \mid \mathbf{U}))=s(\mathbf{U}, \mathbf{y})-\log \left(\sum_{\tilde{\mathbf{y}} \in \mathbf{Y}} e^{s(\mathbf{U}, \tilde{\mathbf{y}})}\right) 
    \label{equa_crf}
\end{equation}

where $Y$ is the set of all possible tag sequences. In the evaluation procedure, Equation \ref{equa_crf} is computed by the Viterbi algorithm \cite{rabiner} to get the maximum-probability label sequence. 

\section{4 Experiments}
In this section, we present the datasets, baselines, metrics and experimental settings.

\subsection{4.1 Datasets}
We evaluate our S+PAGE model on four benchmark datasets -- IEMOCAP, MELD, DailyDialog and EmoryNLP. For this work, we only consider emotion recognition based on textual features. The statistic of them is shown in Table \ref{tab:statistic}.

\subsubsection{IEMOCAP \cite{iemocap}} is a audiovisual dataset. It consists of dyadic conversations where actors perform improvisations or scripted scenarios. Each conversation is segmented into utterances, which are annotated with one of the six emotion labels, which are happy, sad, neutral, angry, excited, and frustrated.

\subsubsection{DailyDialogue \cite{dailydialog}} is a human-written multi-turn dyadic dialogue dataset, reflecting our daily communication way and covering various topics about our daily life. Emotion labels in it contain neutral, happiness, surprise, sadness, anger, disgust, and fear.

\subsubsection{MELD \cite{meld}} is a multi-modal emotion classification dataset. It is a multi-party dialogue dataset created from scripts of the Friends TV series. Each utterance is annotated as one of the seven emotion classes: anger, disgust, sadness, joy, surprise, fear or neutral.

\subsubsection{EmoryNLP \cite{emorynlp}} is also collected from Friends TV series, but it is different from MELD in the choice of scenes and emotion labels. The emotion labels include neutral, sad, mad, scared, powerful, peaceful, and joyful.

For the evaluation metrics, we choose micro-averaged F1 for DailyDialog and weighted-average F1 for the other datasets, following previous works \cite{rgat, dag, ket}.

\begin{center}
\begin{table}[t]
	\centering
	\resizebox{\linewidth}{!}{
	\begin{tabular}{l|c|c|c|c|c|c}
		\toprule
		\multirow{2}*{Dataset} & \multicolumn{3}{c|}{\# Conversations} &\multicolumn{3}{c}{\# Uterrances}\\ 
		&Train&Val&Test&Train&Val&Test\\
		\hline
		IEMOCAP&\multicolumn{2}{c|}{120}&31&\multicolumn{2}{c|}{5810}&1623\\ 
		MELD&1039&114&280&9989&1109&2610\\
		DailyDialog&11118&1000&1000&87170&8069&7740\\
		EmoryNLP&713&99&85&9934&1344&1328\\
		\bottomrule
	\end{tabular}
	}
	\caption{The statistics of four datasets.}
	\label{tab:statistic}
\end{table}
\end{center}

\subsection{4.2 Baseline Methods}
For a comprehensive performance evaluation, We compare S+PAGE with the following baselines:

\subsubsection{CNN:} A single layer convolutional
neural network model.

\subsubsection{RNN-based methods:} CNN+cLSTM \cite{cnn_lstm}, DialogueRNN \cite{dialoguernn}
\subsubsection{KET \cite{ket}:} A transformer structure with hierarchical self-attention and external commonsense knowledge.
\subsubsection{GNN-based methods:} DialogurGCN \cite{dialoguegcn}, RGAT \cite{rgat} and DAG-ERC \cite{dag}.

\subsection{4.3 Experimental Settings}
We set the initial learning rate of 1e-4 in the Transformer layers, 2e-3 in the PAG layers and 2e-2 in the CRF layer. AdamW optimizer is used under a scheduled learning rate following \cite{transformer}. The number of dimensions of the utterance representations and contextual embeddings is set to 300. We set the dropout rate and number of attention head in TSCT to be 0.1 and 8 repectively. 3-head attention is used during calculating the edge weights. We also conduct experiments with different window sizes and PAG layers. We choose the hyper-parameters that achieve the best score on each dataset by using development data. The training and testing process is run on a single Tesla P100 GPU with 32G memory. The reported results of our implemented models are all based on
the average score of 5 random runs on the test sets.
\section{5 Results and Analysis}

\begin{table}[t]
	\centering
	\resizebox{\linewidth}{!}{
	\begin{tabular}{l|c|c|c|c} 
		\toprule
		Model & IEMOCAP &MELD &DailyDialog &EmoryNLP\\ 
		\hline
	    CNN & 48.18 &55.86 &49.34 &32.59\\
	    CNN+cLSTM &54.95 &56.87 &50.24 &32.89\\
	    DialogueRNN &   62.75   &57.03  &-           &-\\
		KET&59.56 &58.18 & 53.37           & 33.95 \\
		\hline
	    DialogueGCN &64.18 & 58.10 &-           &-\\
		RGAT &65.22 &60.91 &54.31          &34.42\\
	    DAG-ERC &68.03 &\textbf{63.65} &59.33 &39.02\\
	    \hline
	    S+PAGE &\textbf{68.72} &63.32 &\textbf{64.07} & \textbf{39.14}\\
		\bottomrule
	\end{tabular}
	}
	\caption{Overall performance on the four datasets. 
	}
	\label{tab:overall}
\end{table}

\subsection{5.1 Overall Performance}
We compare our model with the baseline methods, and the results are reported in Table \ref{tab:overall}. We can note that our proposed S+PAGE has competitive results across the four datasets and achieves a new state-of-the-art performance on the IEMOCAP, DailyDialog and EmoryNLP datasets. 

As shown in the table, all GNN-based models outperform RNN-based models. Compared with existing GNN-based models, our model even has considerable improvement. There are three main advantages that contribute to our performance:  1) contextual modeling with both self and inter-speaker dependency, 2) the presence of relative positional encoding in GNN, 3) consistency modeling of global emotion transfer.


We find that the improvement on MELD and EmoryNLP is not significant. These two datasets are constructed based on Friends TV series, which involve plenty of commonsense knowledge. Thus, pre-trained and knowledge-enhanced models, such as RGAT, DAG-ERC, KET have good results.
However, our model still achieves competitive performance compared to these models as shown in the table.

\subsection{5.2 Ablation Study}

\begin{table}[b]
	\centering
	\resizebox{0.4\textwidth}{!}{
	\begin{tabular}{l|cc}
		\toprule
		Method  &IEMOCAP & MELD\\
		\hline
		S+PAGE &68.72 & 63.32 \\
		\ \ \ \ - TSCT & 67.83 ($\downarrow$0.89) &  61.85 ($\downarrow$1.47) \\ 
		\ \ \ \ - PAG & 64.14 ($\downarrow$4.58) &  61.07 ($\downarrow$2.25) \\
		\ \ \ \ - CRF & 68.07 ($\downarrow$0.65) &  63.01 ($\downarrow$0.31) \\ 
		\bottomrule
	\end{tabular}
	}
	\caption{Results of ablation study.}
	\label{tab:ablation}
\end{table}

For better understanding the contribution of each component in our proposed model, we conduct experiments by replacing TSCT with the vanilla Transformer, and removing PAG and CRF from our model respectively. The results on IEMOCAP and MELD are shown in table \ref{tab:ablation}. 
We can observe that when TSCT is removed, the weighted F1 score drop more on MELD than that on IEMOCAP.
This shows the superiority of TSCT on contextual feature extraction of multi-party conversations, as there are more speakers in dialogues of MELD.
Removal of PAG leads to significant drop on both datasets, which implies the importance of PAG to refine the contextual features with speaker dependency and temporal information.
Meanwhile, after removing CRF layer, we observe the performance degradation. It implies that the modeling of label consistency is essential in the ERC task.
To sum up, all of the three components contribute to the performance improvement of S+PAGE.

\subsection{5.3 PAG Model Depth }

\begin{table}[b]
	\centering
	\resizebox{0.4\textwidth}{!}{
	\begin{tabular}{l|c}
		\toprule
		Method  &IEMOCAP\\
		\hline
		S+PAGE & 68.72\\
	    S+PAGE(-PAG) + GCN  & 64.83 \\ 
	    S+PAGE(-PAG) + RGAT  & 65.62 \\
		S+PAGE(-PAG) + DAG  & 63.48 \\
		\bottomrule
	\end{tabular}
	}
	\caption{Results of replacing PAG with other graph structures.}
	\label{tab:structure}
\end{table}

\begin{figure}[t]
\centering
\includegraphics[width=0.85\columnwidth]{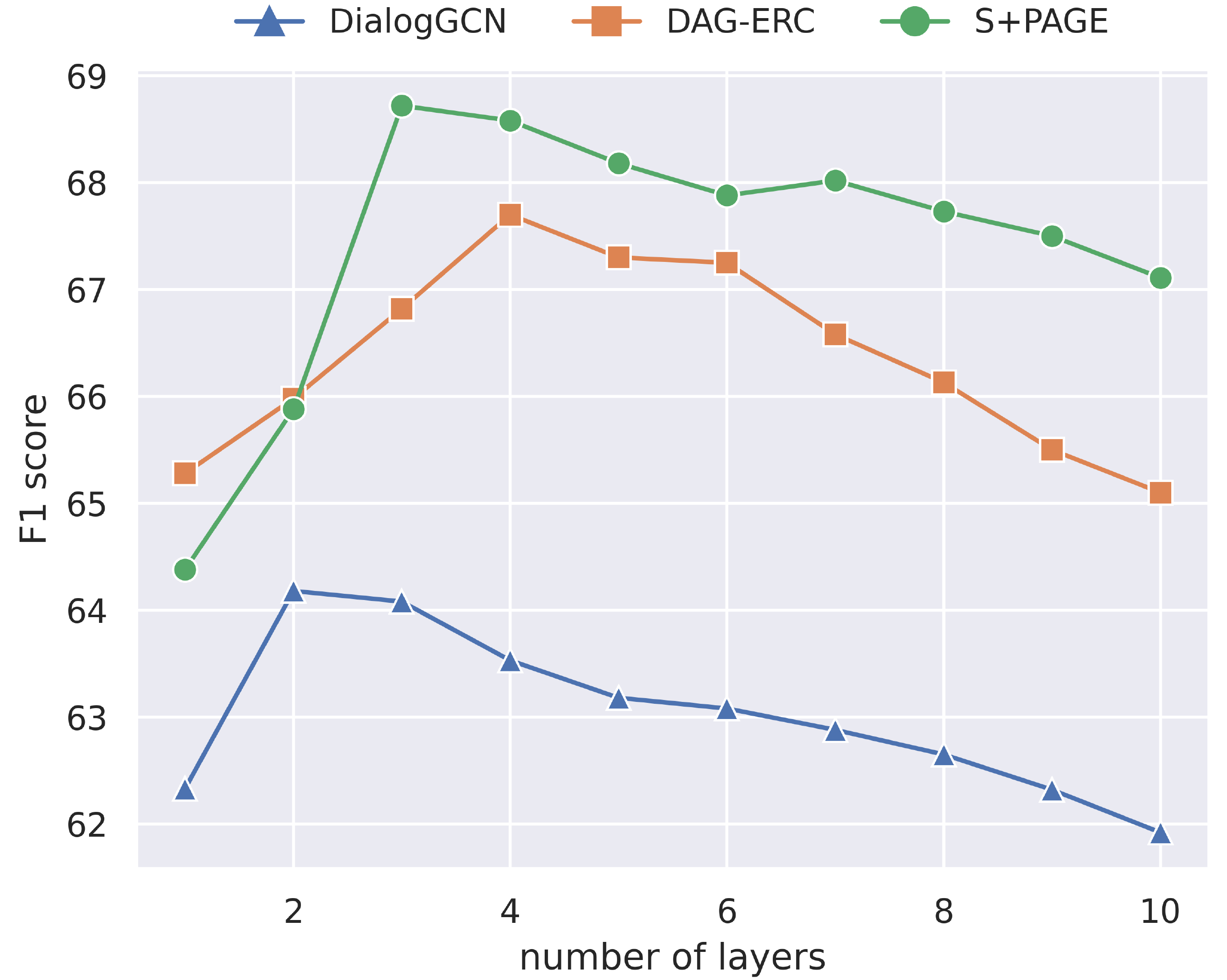} 
\caption{Results of varying depths of GNN.}
\label{fig5}
\end{figure}

We further explore the relationship between model performance and the depth of PAG.
Stacking too many layers of GNN leads to performance degradation due to the over-smoothing problems \cite{over-smooth}. 
As shown in Figure \ref{fig5}, we set different number of layers of PAG on IEMOCAP, and compare the performance with DiaglogGCN and DAG-ERC. 
As can be noted from Figure \ref{fig5}, all the models suffer from performance degradation when the model depth grows. However, the curve of PAG's F1 score descends more slowly than the other methods, which indicates the robustness of our model.

\subsection{5.4 Whether PAG outperforms other graph structures?}
We conduct experiments on IEMOCAP by replacing PAG with the graph structures in DialogueGCN, RGAT and DAG-ERC respectively. 
As shown in Table \ref{tab:structure}, S+PAGE still outperforms the other methods significantly. 
Notice that both DialogueGCN and RGAT with our contextual and consistency modeling perform better than their original versions. It proves the effectiveness of our none-graph parts.
In addition, the performance declines when PAG is replaced by the DAG in DAG-ERC. 
We think the main reason is that we do not use RoBERTa \cite{roberta} in our contextual modeling compared with DAG-ERC.

\subsection{5.5 Effect of Window Size}

\begin{figure}[t]
\centering
\includegraphics[width=0.85\columnwidth]{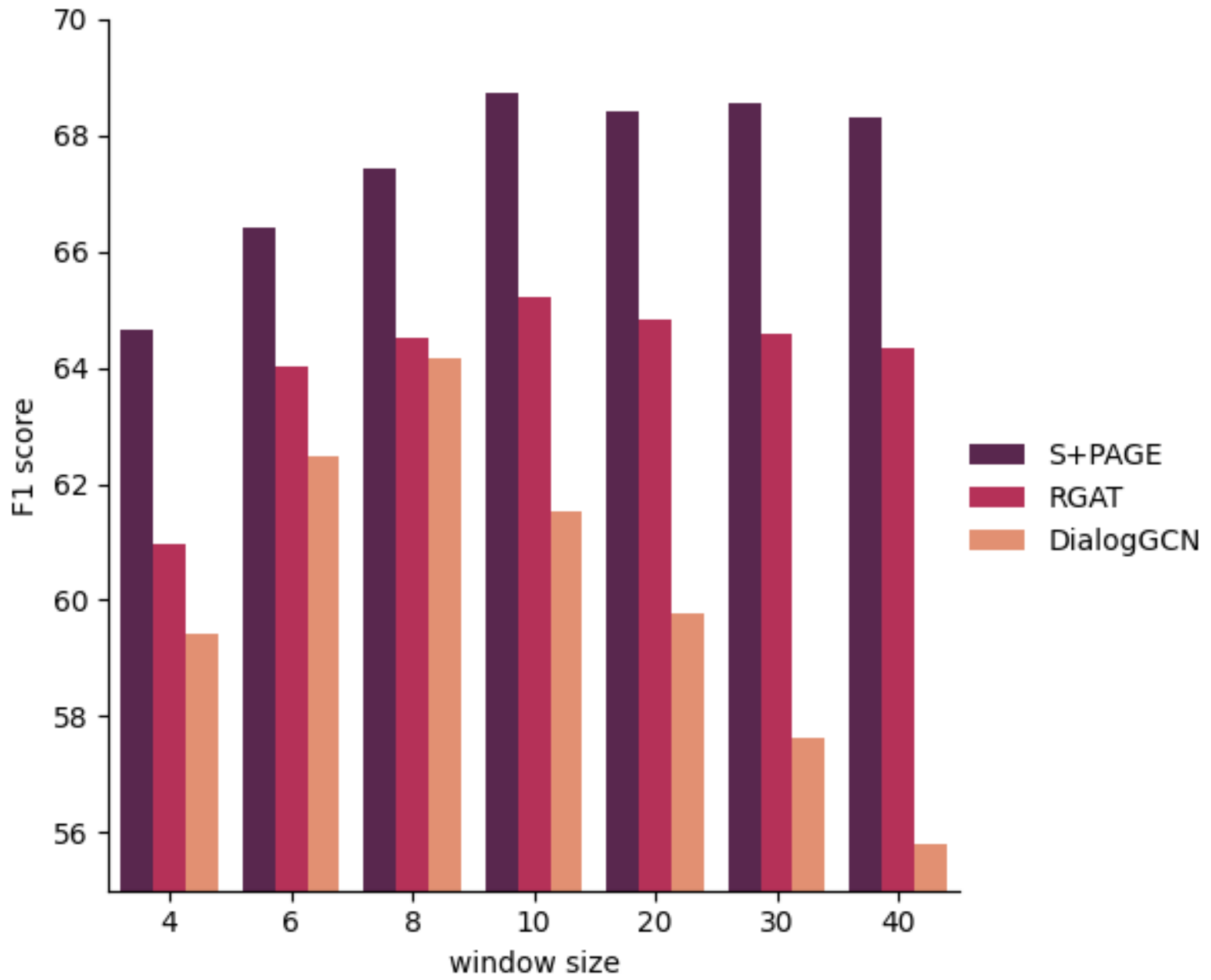} 
\caption{Results of varying window sizes.}
\label{fig6}
\end{figure}

We analyze the influence of past and future window sizes by conducting
experiments with window size $w$ of $(4, 4)$, $(6, 6)$, $(8, 8)$, $(10, 10)$, $(20, 20)$, $(30, 30)$, $(40, 40)$ on IEMOCAP dataset. As shown in Figure \ref{fig6}, the F1 score of S+PAGE, RGAT and DialogueGCN significantly increase,
when the window sizes expand from $4$ to $10$.
The reason is that useful contextual information keeps growing with the increasing of $w$.
However, after window sizes exceed $20$, the F1 score drops for both DialogueGCN and RGAT.
The reason is that the amount of useless long-range dependency increases when the window size continuously grows, which hinders the models from efficiently capturing crucial context.
In contrast, the performance of S+PAGE fluctuates in a relatively narrow range, which shows the robustness of our model on varied window sizes. 
We can infer that the relative positional encoding endows capability of distinguishing critical contextual information to our model.

\section{Conclusion}
In this paper, we propose a novel graph neural network-based model, named S+PAGE, for emotion recognition in conversation (ERC). Specifically, S+PAGE contains three parts, i.e., contextual modeling, speaker dependency modeling, consistency modeling, to incorporate all kinds of contextual information. We present a new Transformer structure with two-stream attention mechanism to better capture the self and inter-speaker conversational context. In speaker dependency modeling, we introduce a novel GNN model, named PAG, to get fine-grained temporal information guided by the relative positional encoding. Furthermore, we use a CRF layer to model emotion transfer in the consistency modeling part. Experiments demonstrate that our model achieves competitive results on several ERC benchmarks. The effectiveness of the proposed two-stream conversational Transformer and position-aware graph neural network is also proved by extensive ablation study.

\bibliography{aaai22.bib}




\end{document}